\documentclass[conference]{IEEEtran}
\IEEEoverridecommandlockouts
\usepackage{cite}
\usepackage{amsmath,amssymb,amsfonts}
\usepackage{algorithmic}
\usepackage{graphicx}
\usepackage{textcomp}
\usepackage{xcolor}
\usepackage{booktabs}
\def\BibTeX{{\rm B\kern-.05em{\sc i\kern-.025em b}\kern-.08em
    T\kern-.1667em\lower.7ex\hbox{E}\kern-.125emX}}
\begin{document}

\title{A Cross-Font Image Retrieval Network for Recognizing Undeciphered Oracle Bone Inscriptions}


\author{\IEEEauthorblockN{Zhicong Wu$^{1,3\dag}$, Qifeng Su$^{2,3\dag}$, Ke Gu$^{1,3}$, Xiaodong Shi$^{2,3*}$}
\IEEEauthorblockA{
\textit{$^1$Institute of Artificial Intelligence, Xiamen University} \\
\textit{$^2$School of Informatics, Xiamen University} \\
\textit{$^3$Key Laboratory of Digital Protection and Intelligent Processing of Intangible Cultural Heritage of }\\
\textit{Fujian and Taiwan (Xiamen University), Ministry of Culture and Tourism}\\
\{zhicongwu, suqifeng, gukexmu\}@stu.xmu.edu.cn, mandel@xmu.edu.cn}
}

\maketitle

\begin{abstract}
Oracle Bone Inscription (OBI) is the earliest mature writing system in China, which represents a crucial stage in the development of hieroglyphs.
Nevertheless, the substantial quantity of undeciphered OBI characters remains a significant challenge for scholars, while conventional methods of ancient script research are both time-consuming and labor-intensive.
In this paper, we propose a cross-font image retrieval network (CFIRN) to decipher OBI characters by establishing associations between OBI characters and other script forms, simulating the interpretive behavior of paleography scholars.
Concretely, our network employs a siamese framework to extract deep features from character images of various fonts, fully exploring structure clues with different resolutions by multiscale feature integration (MFI) module and multiscale refinement classifier (MRC).
Extensive experiments on three challenging cross-font image retrieval datasets demonstrate that, given undeciphered OBI characters, our CFIRN can effectively achieve accurate matches with characters from other gallery fonts, thereby facilitating the deciphering.
\end{abstract}


\section{Introduction}
\label{sec:intro}

\begin{figure}[t]
\centering
\includegraphics[width=0.48\textwidth]{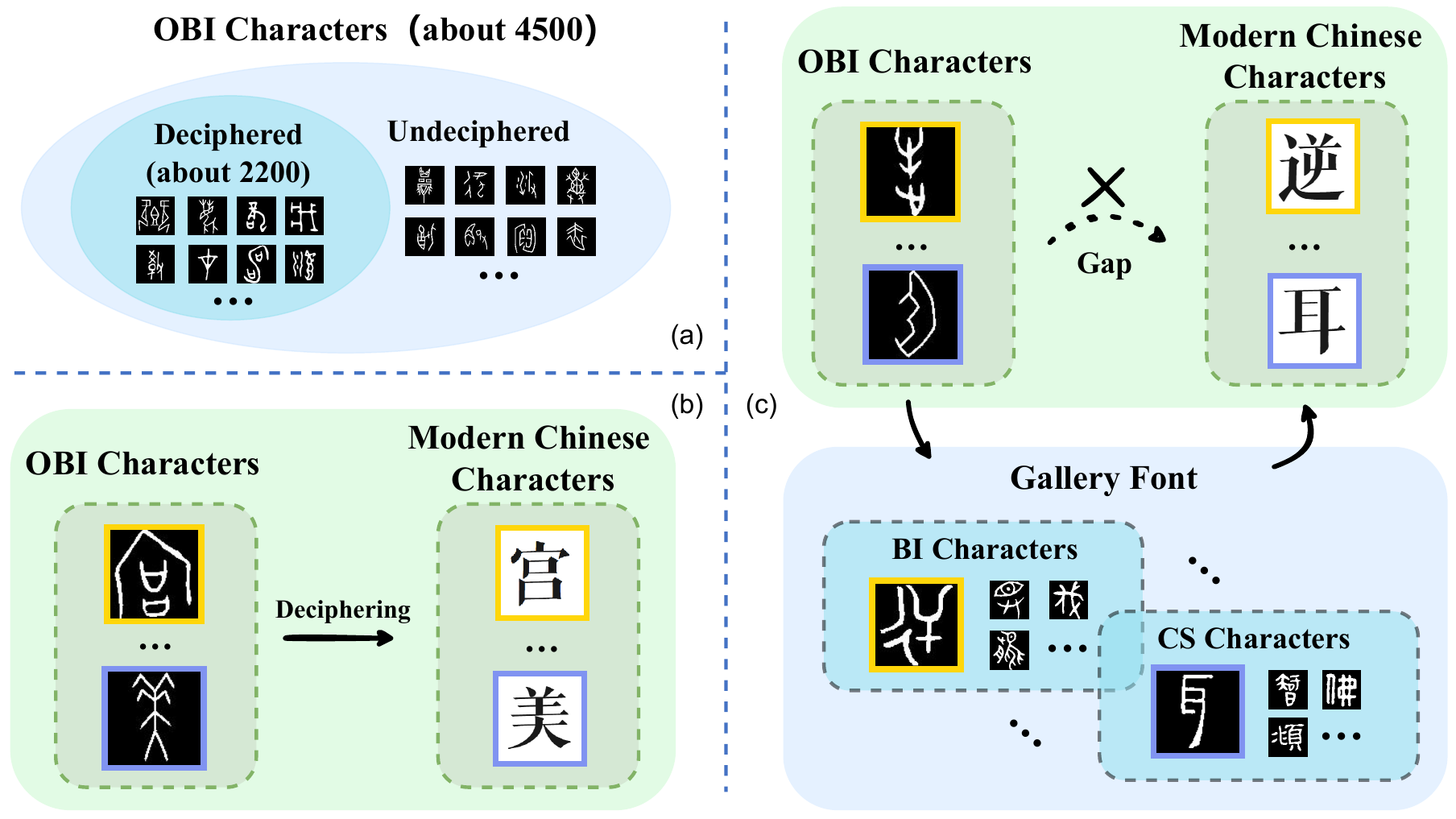} 
\caption{(a) Over half of OBI characters remain undeciphered. (b) The goal of deciphering is to match OBI characters to corresponding modern Chinese characters. (c) Utilizing historical font intermediaries to overcome glyph gaps and enable effective decipherment of undeciphered OBI characters.}
\label{workflow}
\end{figure}

Oracle Bone Inscription (OBI) derives its name from being carved on tortoise shells and animal bones, serving as significant records for divination and military warfare during China's Shang Dynasty.
Therefore, deciphering OBI characters enables scholars to access valuable information recorded thousands of years ago, offering crucial insights into the social structure and cultural practices of early China.
Specifically, as illustrated in Fig. \ref{workflow}(b), the process of deciphering involves identifying and correlating OBI characters with their corresponding modern Chinese counterparts.
To decode undeciphered OBI characters, paleographers require not only extensive linguistic expertise and practical experience but also a deep understanding of the historical development and diverse variations of these characters \cite{takashima1978decipherment}.
Due to the labor-intensive and time-consuming nature of the interpretation process, only approximately 2,200 of the nearly 4,500 OBI characters discovered have been deciphered to date \cite{li2020hwobc} as shown in Fig. \ref{workflow}(a).

With significant advancements in artificial intelligence technology in recent years, an increasing number of researchers have applied advanced AI methods to address various practical challenges in the OBI domain, such as oracle bone character fragments rejoining \cite{jin2023interactively, yuan2023sff}, OBI character detection \cite{liu2020spatial, zhen2024oracle, fu2024detecting, weng2024oracle, huang2024research}, and OBI character recognition \cite{meng2018recognition,han2020self,zhao2020oracle,ge2021oracle,sun2020dual,gao2020distinguishing,wang2022unsupervised,li2024research,hu2024component}, to name a few.

\begin{figure*}[t]
\centering
\includegraphics[width=0.9\textwidth]{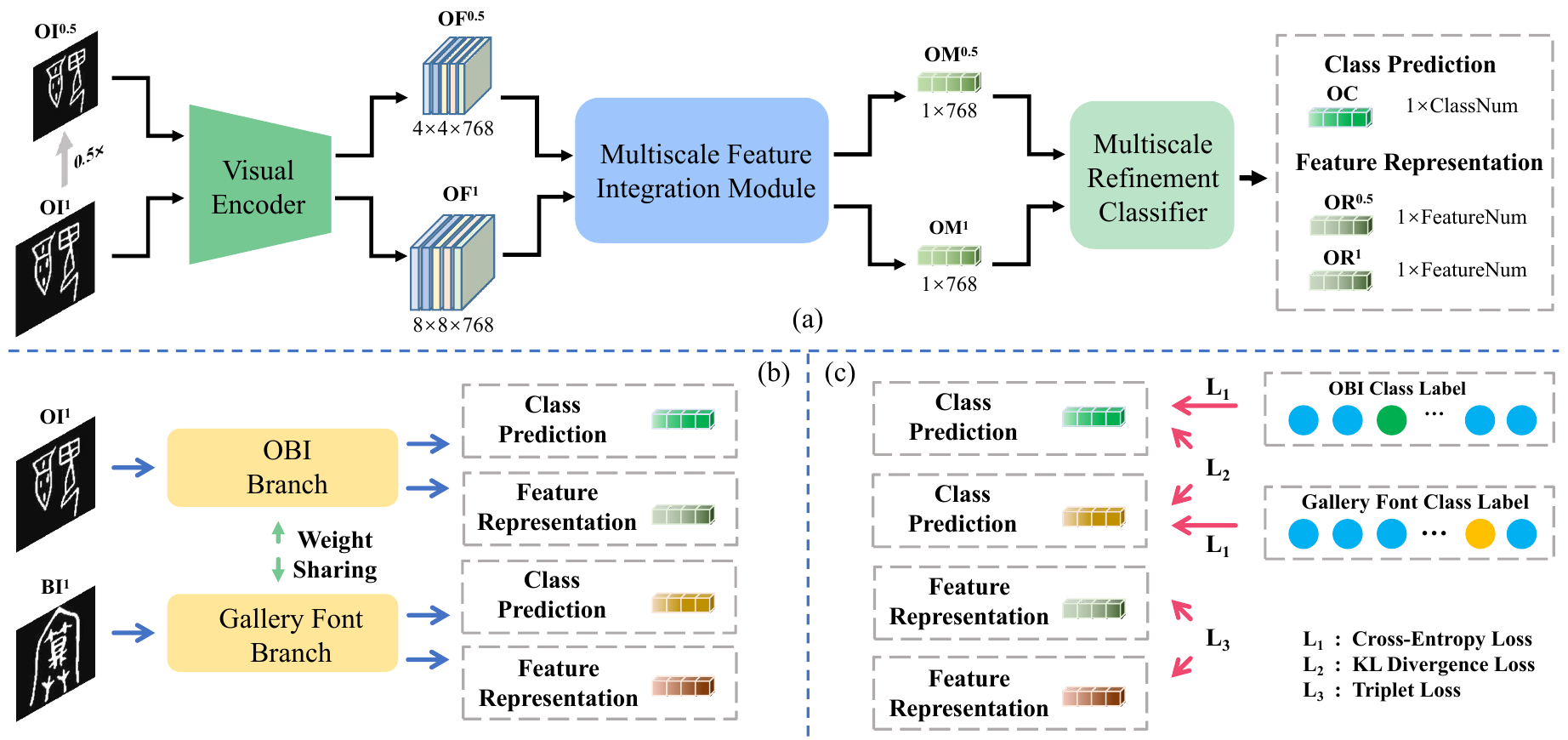} 
\caption{(a) Taking the OBI Branch as an example, high-quality feature extraction and class prediction are performed for the input image $\mathbf{OI^{1}}$. (b) The siamese network architecture of the proposed CFIRN. (c) The loss functions utilized in CFIRN include Cross-Entropy for improving classification accuracy, KL Divergence for aligning feature distributions, and Triplet Loss for optimizing the feature space.}
\label{network}
\end{figure*}

In the field of OBI character decipherment, significant progress have been made through various deep learning-driven approaches. 
Wang et al.\cite{wang2022study} analyzed the evolution process using few-shot learning, while others have explored classification approaches to associate undeciphered OBI with ancient scripts such as the large seal script \cite{xu2023conf}.
The use of diffusion models has also been emerged as a promising avenue for evolving OBI characters into modern forms \cite{li2023diff, guan2024deciphering}. Furthermore, innovative methods involving the disassembly and reconstruction of OBI characters into modern substitutes have been proposed \cite{wang2024puzzle}. 
Gao et al.\cite{gao2024linking} proposed a retrieval-based method to establish a mapping relationship between undeciphered OBI characters and deciphered ones.
Collectively, these studies underscore the potential of deep learning in advancing the automated decipherment. 

However, the significant evolution of character forms throughout an extensive historical period presents an impediment to establishing direct and immediate correlations.
In the concrete practice of paleography, researchers often employ comparative analysis of OBI characters with scripts from various historical periods, particularly those from closely related epochs, to choose more plausible interpretations.

Drawing insights from the methodology of paleographic research, we designed a workflow that utilizes gallery fonts from other historical periods as intermediaries to decipher the undeciphered OBI characters, as shown in Fig. \ref{workflow}(c).
Concretely, we propose a cross-font image retrieval network (CFIRN) that establishes associations between OBI characters and other script forms, with the process depicted in Fig. \ref{network}.
To address the dual challenge in ancient script images, where large-scale inputs emphasize intricate details and small-scale inputs capture broader structural patterns for complementary analysis, we seamlessly combine visual encoder features through the application of multiscale feature integration (MFI) module, further enhancing them with multiscale refinement classifier (MRC) to achieve the generation of highly refined features.
Consequently, three richly annotated ancient scripts, including Bronze Inscription (BI), Bamboo Slip Inscription (BSI) and Clerical Script (CS), are selected as our gallery set.
These scripts are essential due to their historical proximity to OBI and their role in capturing the structural transitions that lead to modern Chinese characters.
Additionally, we conducted exploratory experiments on real-world undeciphered OBI data, where qualitative results further demonstrated the effectiveness of our model.

In summary, our main contributions can be summarized as follows: 1) To the best of our knowledge, this is the first study to explore the capability of a unified cross-font image retrieval network for the task of OBI character decipherment. 2) Our CFIRN framework introduces a novel multiscale feature integration (MFI) module and a multiscale refinement classifier (MRC), to enhance multiscale feature extraction and fusion, enabling robust representation of ancient script characters. 3) Compared to state-of-the-art image retrieval models, our CFIRN demonstrates superior performance on three ancient script image retrieval datasets.

\section{METHOD}
The architecture of the proposed Cross-Font Image Retrieval Network (CFIRN), as shown in Fig. \ref{network}, leverages a Siamese network structure to integrate multiscale deep features extracted from two fonts, which enhances the quality of feature representation, simulating the interpretive behavior of comparing ancient inscriptions with known script forms.
Both the OBI branch and the Gallery Font branch share the same framework, and for clarity, we describe the OBI branch as an example in Sections \ref{Multiscale Feature Integration} and \ref{Multiscale Refinement Classifier}.

\subsection{Feature Extraction Process}

Drawing inspiration from related research \cite{shen2023mccg,deuser2023sample4geo}, we adopted ConvNeXt \cite{liu2022convnet} as the backbone of the encoder network.
The encoder weights are shared between the OBI branch and the Gallery Font branch, enabling better generalization in feature extraction under low-data conditions.
Given an input image pair $\{\mathbf{OI^{1}},\mathbf{BI^{1}}\}$ (comprising an OBI character image and a corresponding gallery font image), we generate their scaled versions $\{\mathbf{OI^{0.5}},\mathbf{BI^{0.5}}\}$ using predefined scaling factors, where $\mathbf{OI^{k}}$ and $\mathbf{BI^{k}}$ represent the input images at scale $k$.
These multiscale images $\{\mathbf{OI^{k}},\mathbf{BI^{k}}\}, k\in\{0.5, 1\}$ are then fed into the visual encoder. Using the split-transform-merge strategy, we obtain the deep features $\{\mathbf{OF^{k}},\mathbf{BF^{k}}\}, k\in\{0.5, 1\}$.

\subsection{Multiscale Feature Integration}
\label{Multiscale Feature Integration}

\begin{figure}[t]
\centering
\includegraphics[width=0.35\textwidth]{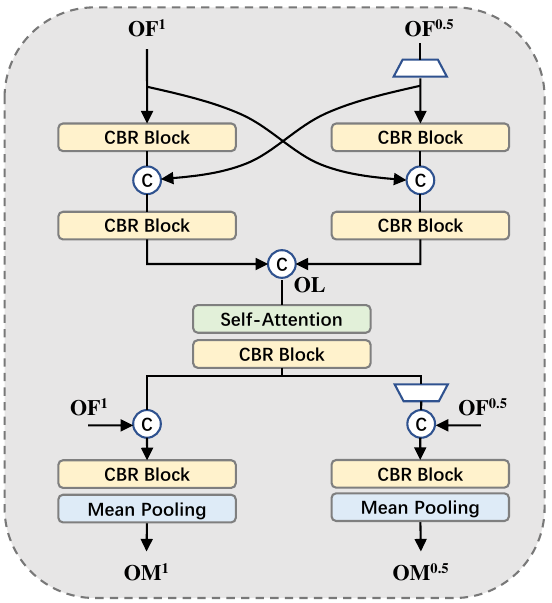} 
\caption{Illustration of the multiscale feature integration (MFI) module.}
\label{MFI}
\end{figure}

The encoder features encompass a wealth of multi-scale information, where the intricate details at various scales intricately complement each other, capturing both fine-grained and coarse structural cues characteristic of ancient characters.
To effectively combine these interactive clues, we propose the Multiscale Feature Integration (MFI) module, as illustrated in Fig. \ref{MFI}.
Following the strategy in \cite{zhou2024decoupling}, we employ interactively placed CBR Blocks, each consisting of a 3×3 convolutional layer, a Batch Normalization (BN) layer, and a ReLU activation function, to fuse the multiscale features. Specifically, the fusion process is defined as follows:
\begin{equation}\label{MFI1}
OL=C(f(C(f(OF^{1}),OF_{\uparrow}^{0.5}),f(C(f(OF_{\uparrow}^{0.5}),OF^{1}),
\end{equation}
where $C(\cdot)$ denotes channel-wise concatenation, $f(\cdot)$ represents the CBR block, and $(\cdot)_{\uparrow}$ indicates a ×2 upsampling operation using bilinear interpolation.

To further refine and combine the fused features, we incorporate the widely-adopted Convolutional Block Attention Module (CBAM), which applies both channel attention and spatial attention mechanisms. This enhances the feature representations by emphasizing important information while suppressing irrelevant details.
Finally, the attention-enhanced features are integrated with the original input features via residual connections, ensuring information preservation. The integrated feature vectors $\{\mathbf{OM^{k}},\mathbf{BM^{k}}\}, k\in\{0.5, 1\}$ are then obtained using mean-pooling layers to produce compact and robust multiscale representations.

\subsection{Multiscale Refinement Classifier}
\label{Multiscale Refinement Classifier}

In the realm of image retrieval tasks, previous studies \cite{lin2022joint, dai2021transformer} have demonstrated that incorporating a classifier can effectively improve retrieval performance.
Here, we propose the Multiscale Refinement Classifier (MRC), which is specifically tailored to refine the integrated features.
The MRC employs Linear Blocks, where each block consists of a linear layer, a Batch Normalization (BN) layer, and a dropout layer. These blocks transform the integrated features into the refined representations $\mathbf{OR^{k}}, k\in\{0.5, 1\}$, where each refined representation has a channel dimension of 512.
The refined features are then passed through the classifier layers in the OBI Branch to produce the fused class prediction $\mathbf{OC}$.
In the testing phase, the classifier is excluded, and the refined representations $\mathbf{OR^{k}}, k\in\{0.5, 1\}$ are concatenated to serve as the final feature representation for OBI characters.
Additionally, we initialize the classifier components using the Kaiming initialization method to ensure stable and efficient training.

\begin{table}[t]
  \centering
  \caption{Overview of Collected Font Image Datasets from Four Historical Periods, Categorized by Modern Chinese Character Classes}
    \begin{tabular}{c|cc}
    \toprule
    Font  & \#images & \#classes \\
    \midrule
    Oracle Bone Inscription (OBI)  & 74994  & 1981  \\
    Bronze Inscription (BI) & 66279  & 3354  \\
    Bamboo Slip Inscription (BSI)  & 18493  & 775  \\
    Clerical Script (CS) & 29596  & 5250  \\
    \bottomrule
    \end{tabular}%
  \label{tab:dataset}%
\end{table}%

\begin{table}[t]
  \scriptsize
  \centering
  \caption{Quantitative comparison of retrieval success rate on the OBI-BI dataset. The best numbers are indicated in bold}
    \begin{tabular}{c|ccccc}
    \toprule
    Method & Recall@1 & Recall@5 & Recall@10 & Recall@1\% & AP \\
    \midrule
    S4G   & 54.54  & 64.38  & 68.36  & 85.01  & 41.20  \\
    RKNet & 64.39  & 73.98  & 76.94  & 89.84  & 51.20  \\
    MCCG  & 64.93  & 75.38  & 78.52  & 90.52  & 49.92  \\
    \midrule
    Ours-R50 & 63.09  & 73.75  & 77.15  & 89.39  & 49.33  \\
    Ours-R101 & 66.65  & 76.25  & 79.11  & 90.48  & 53.23  \\
    \midrule
    Ours-C-T & 69.32  & 78.49  & 81.79  & 91.53  & 55.89  \\
    Ours-C-S & 70.52  & 78.79  & 81.36  & 91.10  & \textbf{58.27} \\
    Ours-C-B & \textbf{70.53} & \textbf{79.08} & \textbf{82.10} & \textbf{91.68} & 58.20  \\
    \bottomrule
    \end{tabular}%
  \label{tab:WithSOTA}%
\end{table}%

\begin{table}[t]
  \scriptsize
  \centering
  \caption{Quantitative result for the proposed CFIRN for different periods.}
    \begin{tabular}{c|ccccc}
    \toprule
    Method & Recall@1 & Recall@5 & Recall@10 & Recall@1\% & AP \\
    \midrule
    OBI→BI & 70.53  & 79.08  & 82.10  & 91.68  & 58.20  \\
    OBI→BSI & 42.68  & 57.72  & 61.95  & 70.78  & 32.52  \\
    OBI→CS & 34.71  & 50.39  & 57.61  & 70.52  & 28.85  \\
    \bottomrule
    \end{tabular}%
  \label{tab:different_periods}%
\end{table}%

\subsection{Loss Function}
Each ancient character is assigned to a corresponding category of modern Chinese characters based on their developmental relationships. As a result, the cross-font image retrieval task is reformulated as an image classification problem, as illustrated in Fig. 2(c).
To optimize the network parameters, we adopt the widely-used cross-entropy loss $CEL_{i}$, which measures the discrepancy between the predicted class distribution and the ground truth. The formulation is as follows:
\begin{equation}\label{eq_CEloss}
CEL_{i}=-\sum_{n=1}^{N}p(x_{i,n})\log q(x_{i,n}),
\end{equation}
where $N$ means the number of classes and $p(x_{i,n})$ and $q(x_{i,n})$ are the ground truth probability and the estimated probability for $i^{th}$ image.

\begin{figure}[t]
\centering
\includegraphics[width=0.45\textwidth]{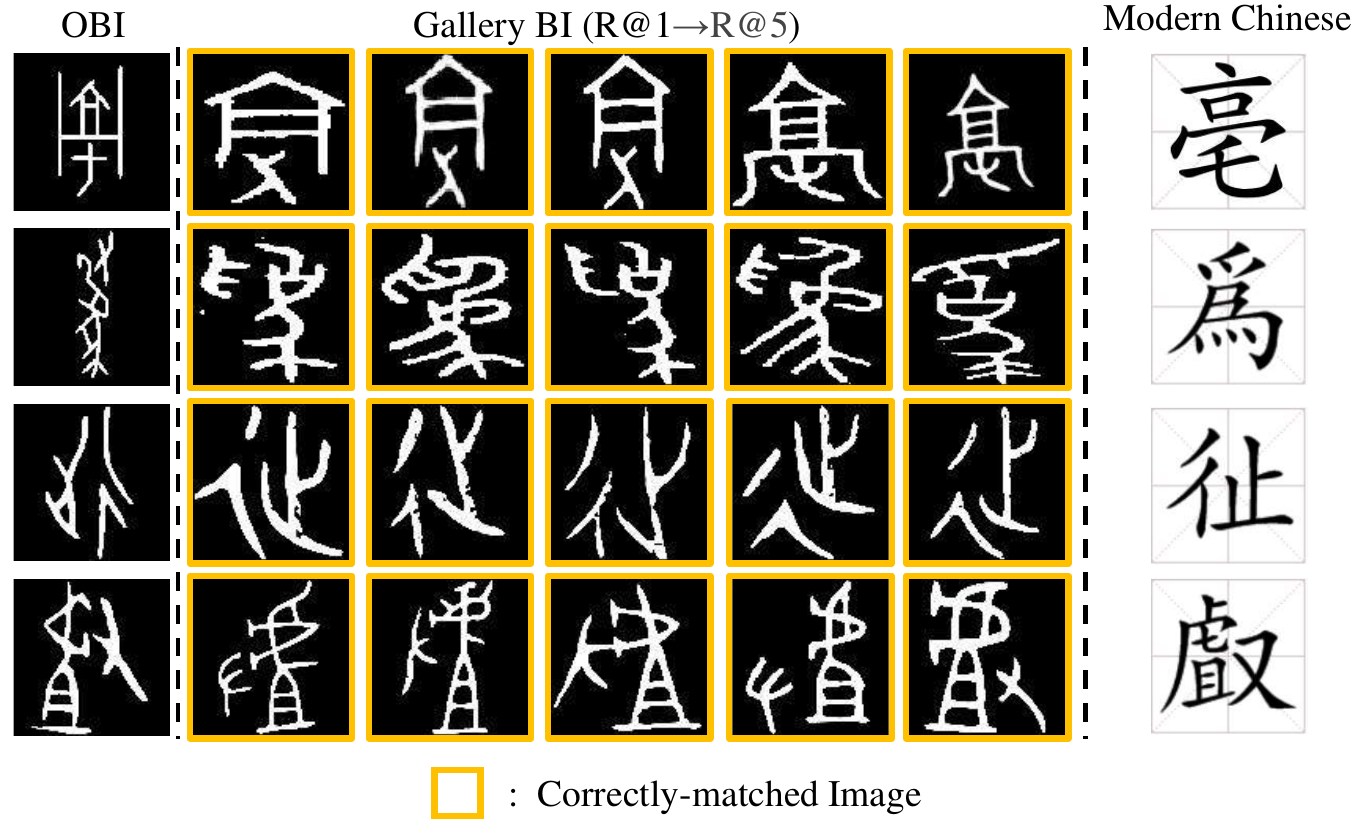} 
\caption{Top-5 retrieval results and corresponding modern Chinese characters on the OBI-BI dataset.}
\label{JW}
\end{figure}

\begin{figure}[t]
\centering
\includegraphics[width=0.4\textwidth]{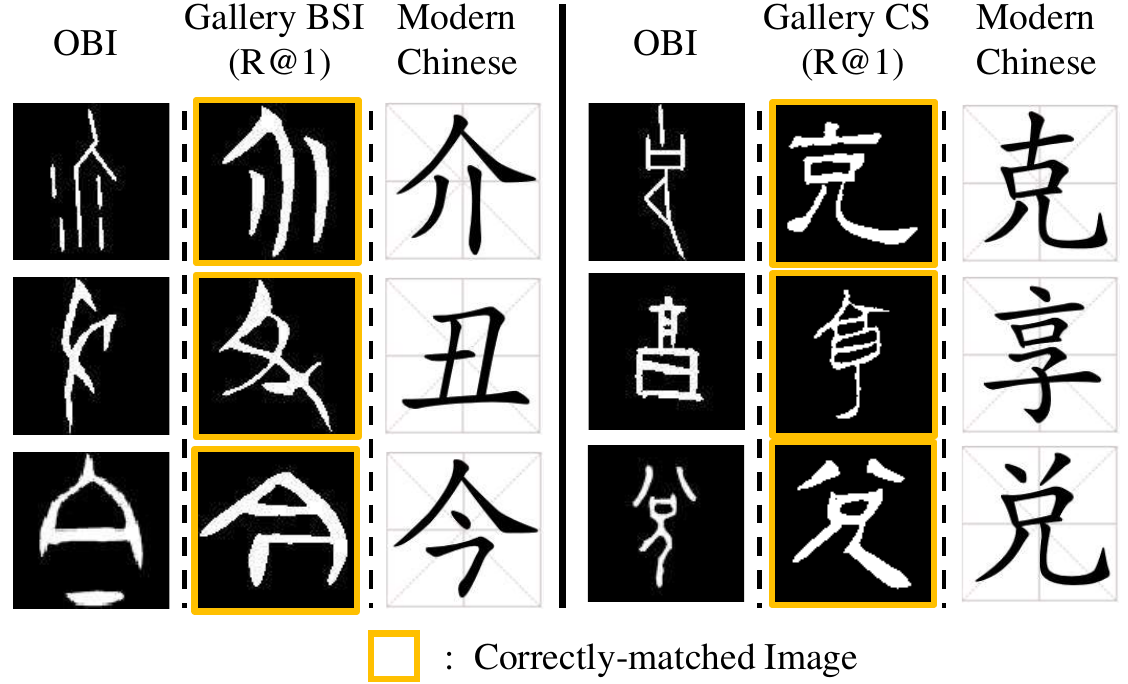} 
\caption{Top-1 retrieval results and corresponding modern Chinese characters on the OBI-BSI dataset and the OBI-CS dataset.}
\label{ZG_LS}
\end{figure}
Through siamese network structure, the input images from the two branches are represented as multi-scale feature vectors.
To align the feature vectors of images from the same category while separating those from different categories, we utilize the triplet loss \cite{chechik2010large}, which is defined as:
\begin{equation}\label{eq_Tloss}
TL_i=max(\|V_i-V_p\|_2-\|V_i-V_n\|_2+M,0),
\end{equation}
where $M$ is the margin value, set to 0.3 in all experiments, and $\|\cdot\|_2$ denotes the $L_2$-distance. Here, $V_i$ is the feature vector of the anchor image, $V_p$ is the feature vector of a positive sample (same category), and $V_n$ is the feature vector of a negative sample (different category).

In addition, to ensure consistency between the predictions of the OBI Branch and the Gallery Font Branch, we employ the Kullback-Leibler (KL) divergence loss. 
The KL loss aligns the probability distributions of the two branches, which helps to improve generalization and stabilize the training process for cross-font retrieval tasks.

In summary, the overall loss function for CFIRN combines three components: cross-entropy loss $CEL$, KL divergence loss $KL$, and triplet loss $TL$.
The total loss $L$ is formulated as::
\begin{equation}\label{eq_loss}
L=CEL+KL+{\alpha}TL,
\end{equation}
where the weight coefficient ${\alpha}$ is set to 5. 
This combination ensures robust feature representation and improved retrieval performance.

\section{EXPERIMENTS}

\subsection{Experiment Setup}

\textbf{Datasets and Evaluation Criteria.}
As shown in Table \ref{tab:dataset}, we collect three ancient script image retrieval datasets for the purpose of OBI character decipherment, sourced from Anyang Normal University and the web.
The gallery font include  Bronze Inscription (BI), Bamboo Slip Inscription (BSI) and Clerical Script (CS).
In the OBI-BI dataset, we conducted experiments with 1358 common character categories with 115655 images, with 30\% of the categories allocated to testing, to simulate real-world challenges.
And the same dataset partitioning strategy is applied to the OBI-BSI dataset and the OBI-CS dataset.
Following studies in image retrieval, we employ several widely used metrics including Recall@K, Recall@Q\%, and average precision (AP) to evaluate the performance of all models.

\textbf{Implementation Details.}
Our proposed CFIRN is implemented with PyTorch, trained end-to-end for 340 epochs with a batch size of 8 on an NVIDIA 2080Ti GPU, and the ConvNeXt pretrained on ImageNet is employed as our vision backbone.
Learning rate initialization is 0.0015 for the backbone parameters and 0.005 for the other layers.
The warm-up period lasts for 80 epochs, after which the learning rate is reduced by a factor of 0.1 at 180 and 240 epochs.
The optimizer chosen is SGD with a momentum of 0.9 and a weight decay of 0.0005.
During the training process, the input images are resized to 256 × 256, and several image augmentation methods are applied, such as random padding, random cropping, and random flipping.

\begin{table}[t]
  \scriptsize
  \centering
  \caption{Ablation study of the components on the OBI-BI dataset. The best numbers are indicated in bold}
    \begin{tabular}{cc|ccccc}
    \toprule
    MRC   & MFI   & Recall@1 & Recall@5 & Recall@10 & Recall@1\% & AP \\
    \midrule
    \checkmark     & \checkmark     & \textbf{69.32} & \textbf{78.49} & \textbf{81.79} & \textbf{91.53} & \textbf{55.89} \\
    \checkmark     &       & 68.39  & 78.10  & 81.51  & 91.36  & 54.94  \\
          &       & 55.86  & 67.05  & 71.32  & 86.32  & 42.36  \\
    \bottomrule
    \end{tabular}%
  \label{tab:ablation}%
\end{table}%

\begin{table}[t]
  \scriptsize
  \centering
  \caption{Ablation study of different resolution strategy on the OBI-BI dataset. The best numbers are indicated in bold}
    \begin{tabular}{c|ccccc}
    \toprule
    Method & Recall@1 & Recall@5 & Recall@10 & Recall@1\% & AP \\
    \midrule
    0.5×, 1.0× & \textbf{69.32} & \textbf{78.49} & \textbf{81.79} & \textbf{91.53} & \textbf{55.89} \\
    1.0×, 2.0× & 68.19  & 77.14  & 80.30  & 90.56  & 55.16  \\
    0.5×, 2.0× & 67.77  & 77.37  & 81.12  & 90.73  & 55.46  \\
    \bottomrule
    \end{tabular}%
  \label{tab:resolution}%
\end{table}%

\subsection{Quantitative Evaluation}
As an image retrieval problem across different domains, we compare our method with the state-of-the-art models, including S4G\cite{deuser2023sample4geo}, MCCG\cite{shen2023mccg} and RKNet\cite{lin2022joint}, in similar cross-view image retrieval tasks.
Table \ref{tab:WithSOTA} highlights the significant advantages of our CFIRN in cross-font image retrieval, showing its superior performance.
Specifically, "Ours-R50/R101" and "Ours-C-T/S/B" denote the backbone architectures of ResNet-50/101 \cite{he2016deep} and ConvNeXt(Tiny/Small/Base), respectively.
"Ours-B" achieves the best performance with a Recall@1 of 70.53\% and an AP of 58.20\%, surpassing the state-of-the-art RKNet by 9.55\% in Recall@1 (from 64.39\% to 70.53\%) and 7.00\% in AP (from 51.20\% to 58.20\%).

In Table \ref{tab:different_periods}, a series of experiments validate the model’s ability to effectively match characters from different historical periods. 
The OBI→BI scenario achieves the best performance with a Recall@1 of 70.53\% and an AP of 58.20\%, significantly outperforming the other two scenarios. This reflects the closer historical proximity and greater structural similarity between Oracle Bone and Bronze Inscriptions, leading to higher retrieval success.
In comparison, the OBI→BSI task achieves moderate performance (Recall@1: 42.68\%, AP: 32.52\%), while the OBI→CS task shows the lowest scores (Recall@1: 34.71\%, AP: 28.85\%),demonstrating the increased difficulty as the time gap widens.
These results align well with real-world paleographic decipherment, where characters from closer historical periods retain more structural similarities, making them easier to associate. 
The findings further validate the robustness of CFIRN in handling cross-font image retrieval tasks across varying levels of character evolution.

\subsection{Qualitative Evaluation}
In Fig \ref{JW} and \ref{ZG_LS}, retrieval results showcase the effectiveness of the CFIRN in matching characters for different historical periods. Our model accurately match Oracle Bone Inscriptions (OBI) with  Bronze Inscription (BI), Bamboo Slip Inscription (BSI) and Clerical Script (CS), and obtain the correlated modern Chinese characters as decipherment. 


The evolution of Chinese characters exhibits diverse morphological variations over time. For instance, certain smaller components may enlarge (Fig. 4, line 1), shift their spatial positions (Fig. 4, line 2), or undergo changes in stroke structure (Fig. 4, line 3). Additionally, the overall character shape may experience left-right flipping (Fig. 4, line 4), further increasing the complexity of character transformations.

As demonstrated by a series of challenging samples, our model successfully retrieves the most similar characters from the gallery datasets, achieving precise matches (highlighted in yellow) across different fonts. These correctly matched results not only reflect the robustness of CFIRN in handling character variations over time but also highlight its ability to bridge the morphological gap between OBI characters and their modern Chinese counterparts.

\subsection{Ablation Studies}
As shown in Tables \ref{tab:ablation} and \ref{tab:resolution}, we conducted comprehensive ablation studies to verify the effectiveness of CFIRN components and configurations.

\textbf{Effectiveness of MFI.}
To evaluate the contribution of the multiscale feature integration (MFI) module to feature extraction, we performed ablation studies as shown in rows 1 and 2 of Table \ref{tab:ablation}.
The results demonstrate that the full model incorporating MFI consistently outperforms the version without it across all metrics, particularly in Recall@1 (69.32\% vs. 68.39\%) and AP (55.89\% vs. 54.94\%).  These results highlight the critical role of MFI in improving model performance.

\textbf{Effectiveness of MRC.}
To isolate and evaluate the contribution of the multiscale refinement classifier (MRC), we compared the performance of models with and without the MRC module (rows 2 and 3 in Table \ref{tab:ablation}).
The results reveal significant performance improvements in Recall@5 (78.10\% vs. 67.05\%) and Recall@10 (81.51\% vs. 71.32\%) when MRC is included.
This improvement arises because the MRC not only refines features but also directly influences the training loss, acting as a crucial component of the model.

\textbf{Mixed-scale Input Scheme.}
 We explored the influence of mixed-scale input combinations on model performance, as detailed in Table \ref{tab:resolution}. 
The results reveal that using both reduced (0.5x) and original (1.0x) input resolutions achieves the best performance, particularly in Recall@1 (69.32\%) and AP (55.89\%), compared to other strategies. 
This suggests that an appropriate mix of scaling factors can effectively enhance model robustness and retrieval precision.

\begin{figure}[t]
\centering
\includegraphics[width=0.45\textwidth]{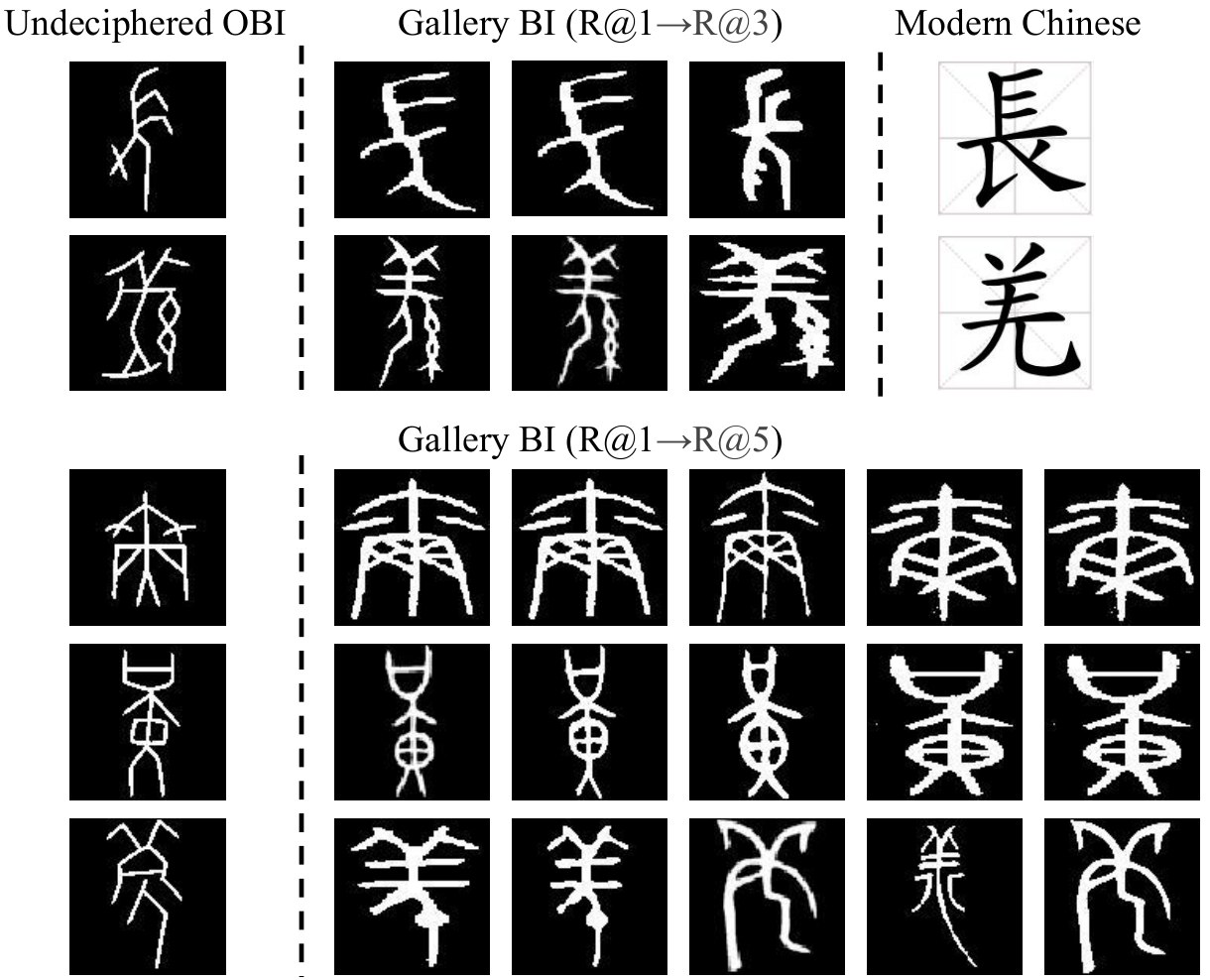} 
\caption{Retrieval results and corresponding modern Chinese characters of the real-world undeciphered OBI characters.}
\label{unknowJGW}
\end{figure}

\subsection{Real-World Undeciphered Character Analysis}

For the decipherment of real-world undeciphered OBI characters, we selected the BI as the gallery font, given its closest temporal proximity and the completeness of its collected data. 
To maximize the utility of limited data resources, we utilized all OBI characters and BI characters that share annotated categories for training. 
In testing, undeciphered OBI characters retrieved all annotated BI characters (3354 classes) to produce the final decipherment results.

As shown in Fig. \ref{unknowJGW}, we illustrate a series of challenging retrieval samples and their corresponding deciphered modern Chinese characters. The retrieval results visually demonstrate the robustness and precision of our model in handling the complex morphological variations inherent in undeciphered OBI characters.
We hope these decipherment results can offer valuable insights for paleographic researchers.

\section{CONCLUSION}
In this article, we propose a novel cross-font image retrieval network (CFIRN), that achieves accurate matching results for OBI character decipherment.
Specifically, we initially harness siamese ConvNeXt-based encoders to extract multiscale features.
Next, to fuse the complementary semantic information, we incorporate the multiscale features from two resolution encoders by deploying the multiscale feature integration (MFI) module.
After that, multiscale refinement classifier (MRC) is employed to further obtain the high quality feature representation, while improve the training effect of the network by classifier block.
Extensive experiments on three challenging cross-font image retrieval datasets demonstrate that our proposed CFIRN can accurately decrypt OBI characters by matching them with characters from other gallery fonts.

\bibliographystyle{IEEEbib}
\bibliography{CFIRN}


\end{document}